\documentclass{article} 
\usepackage{iclr2026_conference,times}

\usepackage{amsmath,amsfonts,bm}

\def\eqref#1{equation~\ref{#1}}

\def\1{\bm{1}}

\DeclareMathAlphabet{\mathsfit}{\encodingdefault}{\sfdefault}{m}{sl}
\SetMathAlphabet{\mathsfit}{bold}{\encodingdefault}{\sfdefault}{bx}{n}

\usepackage{hyperref}
\usepackage{url}
\usepackage{booktabs}
\usepackage{array}
\usepackage{ragged2e}
\usepackage{caption}
\usepackage{graphicx} 
\usepackage{float}

\usepackage{amsmath,amssymb,amsthm}

\title{Emergent Hierarchical Monosemantic Neurons from the Group-Contrastive Forward-Forward Algorithm}

\author{%
  Yiming Tang\thanks{Email: yiming@nus.edu.sg} \\
  National University of Singapore\\
  \And
  Qinglin Qi\\
  Lund University\\
  \And
  Zhaoqian Yao \\
  Chinese University of Hong Kong\\
  \And
  Harshvardhan Saini \\
  Indian Institute of Technology, Dhanbad\\
  \And
  Dianbo Liu\thanks{Corresponding author. Email: dianbo@nus.edu.sg} \\
  National University of Singapore\\
}

\iclrfinalcopy 
\begin{document}

\maketitle

\begin{abstract}
Mechanistic interpretability has made significant strides in understanding neural network representations, with sparse dictionary learning (SDL) methods, most prominently sparse autoencoders, as a central paradigm. However, recent work has reported several limitations of this paradigm: SDL objectives are non-identifiable; SDL methods rely heavily on the Linear Representation Hypothesis; and a growing body of evidence points to concepts that are encoded non-linearly and are therefore not expressible as any single direction. We hypothesise that a different route to monosemanticity is available. Biological visual systems exhibit highly selective neurons organised into hierarchies of increasing abstraction, and this organisation emerges from local, layer-wise learning rules rather than from a global error signal; we therefore ask whether a biologically plausible learning algorithm will likewise yield monosemantic neurons. To test this, we propose Group-Contrastive Forward-Forward (GCFF), a forward-forward training algorithm that combines class-specific routing with within-class contrastive objectives, reaching monosemanticity through architectural constraints rather than sparsity. Because GCFF attaches multiple non-linear layers to the representation under study, its neurons can therefore capture the non-linear concepts. On CLIP representations, a single trained GCFF module recovers monosemantic neurons whose abstraction increases progressively with depth, reaching environmental properties that hold independently of an image's foreground, without any sparsity constraint or supervision of abstraction level. We further demonstrate that GCFF can train networks from scratch, achieving state-of-the-art performance among forward-forward algorithms on various image classification benchmarks. 

\end{abstract}

\section{Introduction}

Understanding the internal representations of neural networks has become one of the central challenges in modern machine learning. As models grow in capability and are deployed in high-stakes domains, the ability to inspect and interpret what concepts a network has learned, and how those concepts are organised, is essential for building trust, diagnosing failures, and advancing our scientific understanding of intelligence itself~\citep{lipton2017mythosmodelinterpretability, rudin2019stopexplainingblackbox}. This need has given rise to the field of mechanistic interpretability~\citep{sharkey2025openproblemsmechanisticinterpretability}, which pursues two central goals: reverse-engineering network computations into human-understandable circuits and algorithms~\citep{olah2020zoom, olsson2022incontextlearninginductionheads}, and identifying the semantic concepts encoded in pretrained representations~\citep{bricken2023monosemanticity, templeton2024scaling, bereska2024mechanisticinterpretabilityaisafety}.

Sparse dictionary learning (SDL) has become the central paradigm for the second of these goals. Sparse autoencoders (SAEs)~\citep{cunningham2023sparseautoencodershighlyinterpretable, bricken2023monosemanticity, gao2024scalingevaluatingsparseautoencoders}, transcoders~\citep{dunefsky2024transcodersinterpretablellmfeature, paulo2025transcodersbeatsparseautoencoders} and crosscoders~\citep{lindsey2024crosscoders} decompose polysemantic neuron activations into individual monosemantic features by imposing sparsity constraints on auxiliary models~\citep{tang2026unifiedtheorysparsedictionary,nelson2026identifiablesparseautoencoders}, and have proven effective at recovering interpretable atomic concepts at scale~\citep{templeton2024scaling}. Recent work has nevertheless reported several limitations of this paradigm. First, SDL objectives are non-identifiable: the training problem is piecewise biconvex and admits spurious minima, so the recovered dictionary is not uniquely determined by the data~\citep{tang2026unifiedtheorysparsedictionary}. Second, every method in the family reconstructs an activation as a sparse \emph{linear} combination of dictionary atoms, which writes the Linear Representation Hypothesis~\citep{park2024linearrepresentationhypothesisgeometry} directly into the decoder: a feature can be recovered only as a direction. Third, a growing body of evidence points to concepts that no single direction expresses, including features that are irreducibly multi-dimensional~\citep{engels2025languagemodelfeaturesonedimensionally} and representations that lie on curved manifolds~\citep{Chung_2021, Sorscher_2022, Cohen2019}. Such concepts remain beyond reach however large the dictionary grows, because the constraint lies in the reconstruction map rather than in the dictionary size.

A further consequence of these dictionary learning approaches is that features are returned as an unstructured set of directions, blind to the hierarchical conceptual structures that organise them. A neuron selective for ``Golden Retriever'' and a neuron selective for ``various dogs'' and a neuron selective for ``animal scenes'' are recovered as peers, with no account of their relationships~\citep{park2025geometrycategoricalhierarchicalconcepts}. When the concepts encoded in a model's representations carry inherent hierarchical structure, methods that recover features as unorganised sets will systematically misrepresent the organisation of knowledge and concept hierarchy, leaving unexplained how networks represent different levels of abstraction~\citep{tang2026unifiedtheorysparsedictionary, engels2025languagemodelfeaturesonedimensionally}.

We argue that these limitations are not incidental but architectural: they follow from the linear reconstruction itself, and no amount of scaling removes them. We therefore ask whether monosemanticity can be reached by a route that does not begin from a linear reconstruction at all. We take inspiration from biological visual and cognitive systems which exhibit clear representational hierarchies, where neurons at successive levels respond to increasingly abstract concepts, with each level composing over the previous~\citep{hubel1962receptive, tanaka1996inferotemporal, quiroga2005invariant}. Crucially, this hierarchy emerges from local, layer-wise learning rules of biological neural circuits---not from global error signals~\citep{lillicrap2020backpropagation}. We hypothesise that biologically plausible training algorithms, which more closely mirror the local learning dynamics of biological circuits, are more likely to produce biologically plausible representations, including the selectivity and emergent hierarchical organisation observed in living systems~\citep{tosato2025emergentrepresentationsnetworkstrained}.

In this work, we propose Group-Contrastive Forward-Forward (GCFF)~\citep{hinton2022forwardforwardalgorithmpreliminaryinvestigations, chen2025selfcontrastiveforwardforwardalgorithm}, a forward-forward training algorithm that offers an alternative path toward monosemanticity through architectural constraints and contrastive learning rather than sparsity. GCFF introduces two key innovations: class-specific routing, where dedicated neuron groups exclusively process samples from their assigned coarse category, forcing intra-class feature discovery; and class-level contrastive learning, which forms positive pairs from distinct samples sharing the same coarse label. Each layer is trained by a local goodness objective, with no gradient propagated across layers. Because GCFF attaches multiple non-linear layers to the representation under study, its neurons are not directions in that representation, and can therefore capture concepts that no single direction expresses.

We apply GCFF as an interpretation module on top of the pretrained representations of a vision-language model, and find that a single trained module produces monosemantic neurons whose abstraction increases progressively with depth: from concrete objects, through functional groupings, to environmental properties that hold independently of an image's foreground. No sparsity constraint is imposed and no supervision of abstraction level is provided; the hierarchy emerges from depth alone. To our knowledge, no prior interpretability method recovers such an ordered hierarchy of monosemantic features without prescribing its levels in advance. Beyond interpretability, GCFF can also train neural networks from scratch, achieving state-of-the-art performance among forward-forward algorithms on image classification benchmarks~\citep{hinton2022forwardforwardalgorithmpreliminaryinvestigations, chen2025selfcontrastiveforwardforwardalgorithm}.

Our contributions are listed as follows:
\begin{itemize}
\item We propose GCFF, a biologically plausible forward-forward training algorithm that serves as an interpretation module for extracting monosemantic features from pretrained representations, without any sparsity constraint.
\item We show that a single GCFF module attached to a vision-language model recovers monosemantic neurons whose abstraction increases progressively with depth.
\item We establish that GCFF, used to train networks from scratch, achieves state-of-the-art performance among forward-forward algorithms on image classification benchmarks.
\end{itemize}

\section{Related Works}
\label{sec:related_works}


\subsection{Concept-based Mechanistic Interpretability}
Mechanistic interpretability aims to reverse-engineer the internal computations of neural networks into human-understandable components~\citep{olah2020zoom, sharkey2025openproblemsmechanisticinterpretability}, and to decode neural representations into human-understandable contents~\citep{zhao2026rep2textdecodingtextsingle}. A central challenge is \textit{polysemanticity}, where individual neurons encode multiple unrelated concepts due to \textit{superposition}e~\citep{elhage2022toymodelssuperposition, park2024linearrepresentationhypothesisgeometry, engels2025languagemodelfeaturesonedimensionally}. To address this, Sparse Dictionary Learning (SDL) methods train auxiliary sparse models to decompose polysemantic activations into monosemantic features~\citep{tang2026unifiedtheorysparsedictionary}, including Sparse Autoencoders (SAEs)~\citep{cunningham2023sparseautoencodershighlyinterpretable, bricken2023monosemanticity}, transcoders~\citep{dunefsky2024transcodersinterpretablellmfeature}, crosscoders~\citep{lindsey2024crosscoders}, and Matryoshka SAEs~\citep{bussmann2025learningmultilevelfeaturesmatryoshka}, with subsequent work scaling these approaches to frontier language models~\citep{templeton2024scaling}. These concept representations have enabled diverse downstream applications, including human-like visual content analysis~\citep{tang2026humanlikecontentanalysisgenerative}, automatic interpretation of physical plausibility failures~\citep{tang2025doesmodelfailautomatic}, interpretable persona control via gradient ascent~\citep{saini2026bridgingmechanisticinterpretabilityprompt}, and enhanced biomedical interpretability~\citep{tang2026cxrlaniclanguagegroundedinterpretableclassifier}.

\subsection{The Forward-Forward Algorithm}
The Forward-Forward (FF) algorithm~\citep{hinton2022forwardforwardalgorithmpreliminaryinvestigations} replaces backpropagation with two forward passes on positive and negative data, where each layer optimizes a local goodness function independently, offering a biologically plausible alternative to gradient-based learning. A key motivation is its suitability for lifelong and on-device learning, as eliminating the backward pass removes the need to store intermediate activations~\citep{chen2025selfcontrastiveforwardforwardalgorithm}. Several variants have since extended FF's capabilities: Layer Collaboration~\citep{lorberbom2023layercollaborationforwardforwardalgorithm} enables inter-layer communication; SymBA~\citep{lee2023symbasymmetricbackpropagationfreecontrastive} introduces symmetric contrastive objectives; the Cascaded Forward Algorithm~\citep{zhao2023cascadedforwardalgorithmneural} improves layer-wise convergence; the Integrated Forward-Forward Algorithm~\citep{tang2023integratedforwardforwardalgorithmintegrating} first combines FF with shallow backpropagation via local losses; and SCFF~\citep{chen2025selfcontrastiveforwardforwardalgorithm} addresses negative sample generation by self-contrasting each input, achieving state-of-the-art among unsupervised local learning algorithms on standard benchmarks. 
\begin{figure}[h]
    \centering
    \includegraphics[width=1\linewidth]{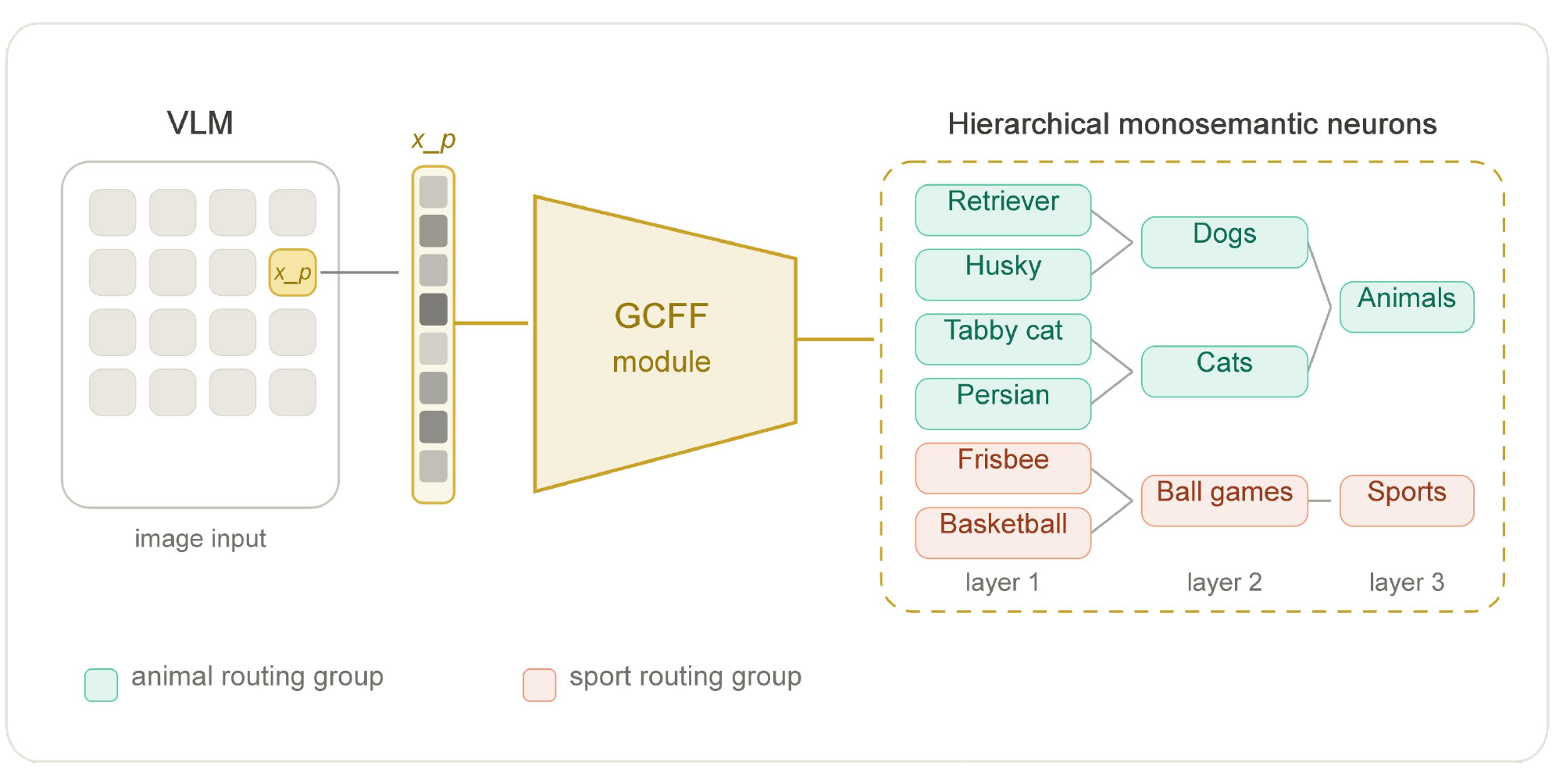}
    \caption{The general framework using the Group-Contrastive Forward-Forward algorithm (GCFF) to interpret the neural representations of a vision-language model.}
    \label{fig:concept_hierarchy}
\end{figure}

\section{The Group-Contrastive Forward-Forward Algorithm}
In this section, we present the Group-Contrastive Forward-Forward algorithm (GCFF), a biologically plausible forward-forward training algorithm that extracts hierarchical monosemantic features from pretrained model representations without sparsity constraints. Section~\ref{sec:overview} gives an overview of the architecture and training pipeline. Section~\ref{sec:pairs} describes the construction of positive and negative sample pairs. Section~\ref{sec:routing} introduces class-specific routing, which enforces neuron specialization within semantic categories. Section~\ref{sec:objective} presents the class-level contrastive learning objective.

\subsection{Overview}
\label{sec:overview}

\begin{figure}
    \centering
    \includegraphics[width=1\linewidth]{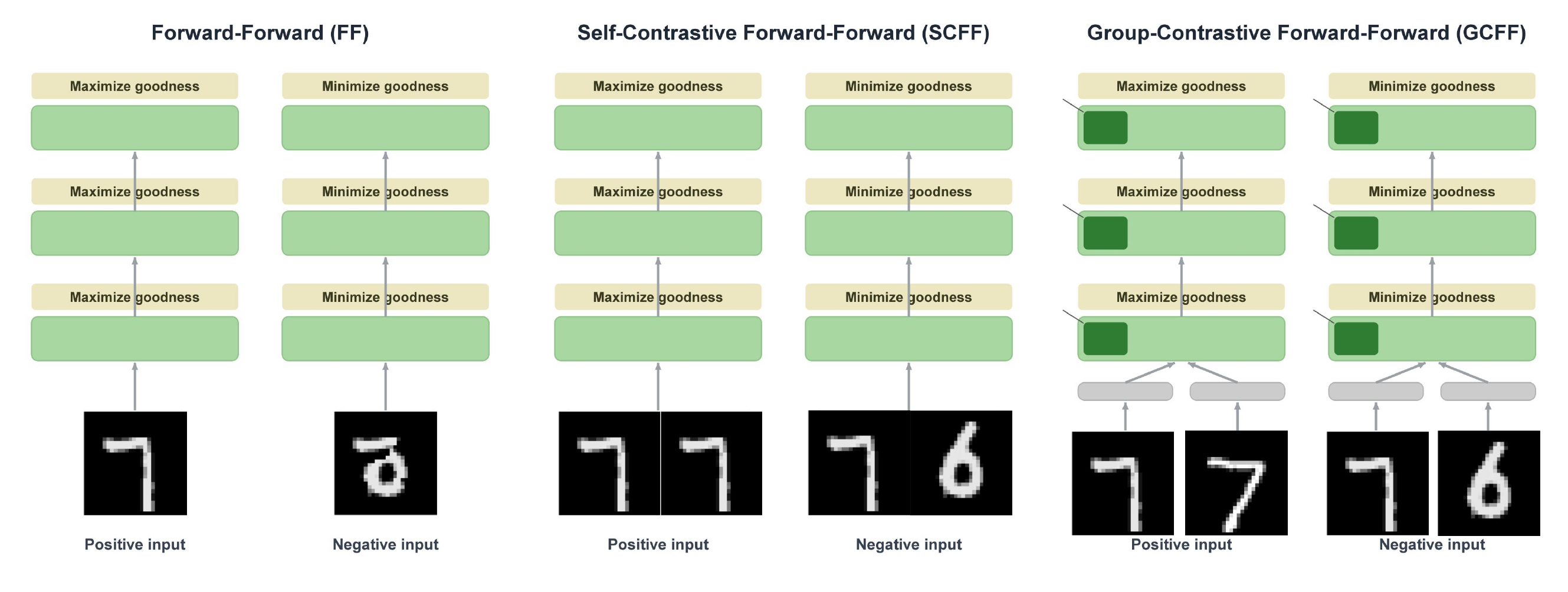}
    \caption{Comparison of Forward-Forward training paradigms. \textbf{FF} (left) uses real vs. corrupted samples. \textbf{SCFF} (middle) concatenates identical samples (positive) vs. different samples (negative). \textbf{GCFF} (right) concatenates same-class samples (positive) vs. different-class samples (negative) and routes inputs through class-specific neuron groups (dark green).}
    \label{fig:method}
\end{figure}

Group-Contrastive Forward-Forward (GCFF) is a forward-forward training algorithm that discovers monosemantic features with emergent hierarchical organization from pretrained model representations (See Figure~\ref{fig:method}). GCFF builds upon the Forward-Forward algorithm~\citep{hinton2022forwardforwardalgorithmpreliminaryinvestigations} and its self-contrastive variant SCFF~\citep{chen2025selfcontrastiveforwardforwardalgorithm}, inheriting their biologically plausible, layer-wise local training dynamics while introducing key innovations that drive hierarchical monosemanticity.

Unlike SDL methods that impose sparsity constraints on auxiliary models to recover monosemantic features post-hoc, GCFF achieves monosemanticity through architectural constraints during training. We train multilayer perceptron (MLP) on top of pretrained representation, using positive and negative sample pairs. The training objective encourages neurons to exhibit high activation for positive pairs (same coarse category) and low activation for negative pairs (different coarse categories).

Two components work in concert to produce hierarchical monosemantic features, including class-specific routing and class-level contrastive learning. Class-specific routing partitions neurons into dedicated groups, each exclusively processing samples from one coarse semantic category, forcing intra-class feature discovery that leads to hierarchical monosemanticity. Class-level contrastive learning forms positive pairs from distinct samples sharing the same coarse label, compelling the network to identify shared fine-grained structure within each category.

\subsection{Positive Samples and Negative Samples}
\label{sec:pairs}

Given a batch of $N$ embeddings $\{\mathbf{x}_i\}_{i=1}^N$ with coarse class labels $\{y_i\}_{i=1}^N$, we generate training pairs using a class-aware selection strategy. For each sample $\mathbf{x}_i$ with coarse label $y_i$, we create $K$ positive pairs by randomly selecting other samples from the \emph{same coarse class}:
\begin{equation}
\mathbf{x}_{\text{pos}} = \mathbf{x}_i + \mathbf{x}_j, \quad \text{where } y_i = y_j \text{ and } i \neq j
\end{equation}
We create $K$ negative pairs by selecting samples from \emph{different coarse classes}:
\begin{equation}
\mathbf{x}_{\text{neg}} = \mathbf{x}_i + \mathbf{x}_k, \quad \text{where } y_i \neq y_k
\end{equation}
By pairing samples from the same coarse category that may represent different fine-grained subcategories, for example, two different dog breeds within the ``canine'' category, the network is ideally encouraged to identify shared intra-class structures.




\subsection{Class-Specific Routing}
\label{sec:routing}

We partition GCFF neurons into class-specific groups, where each group $G_c$ exclusively processes samples from one coarse category $c$ during training. For a layer $l$ with $M^{(l)}$ total neurons serving $C$ coarse classes, we allocate $k^{(l)} = M^{(l)}/C$ neurons per class, choosing layer widths so that $k^{(l)}$ is an integer. An input embedding $\mathbf{x} \in \mathbb{R}^{d}$ with coarse class label $c$ activates only its corresponding neuron group, and the routing decision persists across all $L$ layers of the module:
\begin{equation}
\mathbf{h}_c^{(1)} = \mathrm{ReLU}\!\left(\mathbf{W}_c^{(1)} \mathbf{x}\right), \qquad
\mathbf{h}_c^{(l)} = \mathrm{ReLU}\!\left(\mathbf{W}_c^{(l)} \mathbf{h}_c^{(l-1)}\right), \quad l = 2, \dots, L,
\end{equation}
where $\mathbf{W}_c^{(1)} \in \mathbb{R}^{k^{(1)} \times d}$ and $\mathbf{W}_c^{(l)} \in \mathbb{R}^{k^{(l)} \times k^{(l-1)}}$ are the weights specific to class $c$. A GCFF module is therefore a set of $C$ parallel $L$-layer networks that share no weights, each trained by the local objective of Section~\ref{sec:objective}.

This architectural constraint ensures that each neuron group exclusively observes samples from its assigned semantic category during training. Since these neurons never encounter samples from other categories, they cannot learn inter-class boundaries. Instead, they are forced to discover intra-class distinctions: the fine-grained subcategory structure within each coarse category, such as the breeds that populate a ``canine'' category rather than what separates canines from vehicles.

\subsection{Class-Level Contrastive Learning}
\label{sec:objective}

For each layer $l$, we define a goodness function that measures the mean squared activation of the active neuron group:
\begin{equation}
G^{(l)}(\mathbf{h}) = \frac{1}{M^{(l)}} \sum_{m=1}^{M^{(l)}} (h_m^{(l)})^2
\end{equation}
where $M^{(l)}$ is the number of active neurons in layer $l$, i.e., the $k$ neurons in the class-specific group being processed. The training objective encourages high goodness for positive pairs and low goodness for negative pairs:
\begin{equation}
\mathcal{L}^{(l)} = -\mathbb{E}_{\mathbf{x}_{\text{pos}}} \log \sigma(G^{(l)}(\mathbf{x}_{\text{pos}}) - \theta_{\text{pos}}) - \mathbb{E}_{\mathbf{x}_{\text{neg}}} \log \sigma(\theta_{\text{neg}} - G^{(l)}(\mathbf{x}_{\text{neg}}))
\end{equation}
where $\sigma$ is the sigmoid function and $\theta_{\text{pos}}, \theta_{\text{neg}}$ are threshold hyperparameters. We implement this using identical weight matrices $\mathbf{W}_c$ for both elements of each pair~\citep{chen2025selfcontrastiveforwardforwardalgorithm}, which is computationally equivalent to summing the inputs before applying the transformation. Each layer is trained independently using this local objective, with no gradient signal propagated across layers.

\section{Experiments}

In this section we present our experimental results. Section~\ref{sec:exp_setup} outlines the experimental setup. Section~\ref{sec:eval_monosemanticity} quantifies feature monosemanticity using an automated LLM evaluation pipeline. Section~\ref{sec:emergent_hierarchy} analyses the emergent cross-layer hierarchical composition of the discovered features. Section~\ref{sec:classification} establishes GCFF's state-of-the-art classification performance among forward-forward algorithms, confirming that interpretability does not compromise task competence.

\subsection{Experimental Setup}
\label{sec:exp_setup}

We evaluate the Group-Contrastive Forward-Forward (GCFF) algorithm as an interpretation module on the representations of a vision-language model. We extract embeddings using CLIP-ViT-L/14 on two standard datasets: ImageNet-1K \citep{deng2009imagenet} and STL-10 \citep{coates2011stl10}. For each dataset we train a three-layer GCFF module.

Class-specific routing requires a coarse partition of the label space. For STL-10 we use its \textbf{10} classes directly; for ImageNet-1K we group the 1000 fine-grained classes into around 300 coarse categories via the coarse categories available in the dataset. With $M^{(l)}$ neurons in layer $l$ serving $C$ coarse categories, each category is allocated $k^{(l)} = M^{(l)}/C$ neurons.

\subsection{Quantitative Evaluation of Monosemanticity via LLM Interpretation}
\label{sec:eval_monosemanticity}

To test whether GCFF can discover genuinely monosemantic neurons, we established an automated evaluation pipeline utilising Large Language Models (LLMs). We computed each neuron's response to every image in the evaluation set, ranked the images, and retrieved the top-20 response images. We calculated each neuron's internal consistency score using the average pairwise cosine similarity of the CLIP embeddings of these top-20 images to screen for candidate interpretable neurons. The top 20 neurons per layer were selected for further evaluation.

We prompted GPT-5.4-mini to generate structured, holistic captions for each top-20 image, covering objects, central focus, colour, and scene. GPT-5.4-mini then generated a single hypothetical explanation sentence capturing the salient, recurring semantic concept across the images. Finally, we presented this explanation and the image captions to two independent evaluators (GPT-5.4-mini and Gemini-3-Flash), prompting them to score the presence of the explanation as 0 (absent) or 1 (present). The proportion of ``1'' scores serves as the alignment score, indicating the neuron's monosemanticity.

Overall, the learned GCFF neurons were highly explainable and selective, exceeding the 0.5 meaningfulness threshold established in similar pipelines \citep{wasserman_2025_brainexplore}. The two evaluators also showed high consistency in their judgments (Table~\ref{tab:monosemanticity}).

\begin{table}[h]
\centering
\caption{Monosemanticity evaluation results across 430 selected neurons (8,600 scored images).}
\label{tab:monosemanticity}
\begin{tabular}{lc}
\toprule
Metric & Value \\
\midrule
Mean alignment score (GPT-5.4-mini) & 0.829 \\
Mean alignment score (Gemini-3-Flash) & 0.824 \\
Exact agreement between evaluators & 91.0\% \\
Cohen's $\kappa$ & 0.687 \\
\bottomrule
\end{tabular}
\end{table}

\subsection{Analysis of Emergent Feature Hierarchy}
\label{sec:emergent_hierarchy}

\begin{figure}[h]
    \centering
    \includegraphics[width=1\linewidth]{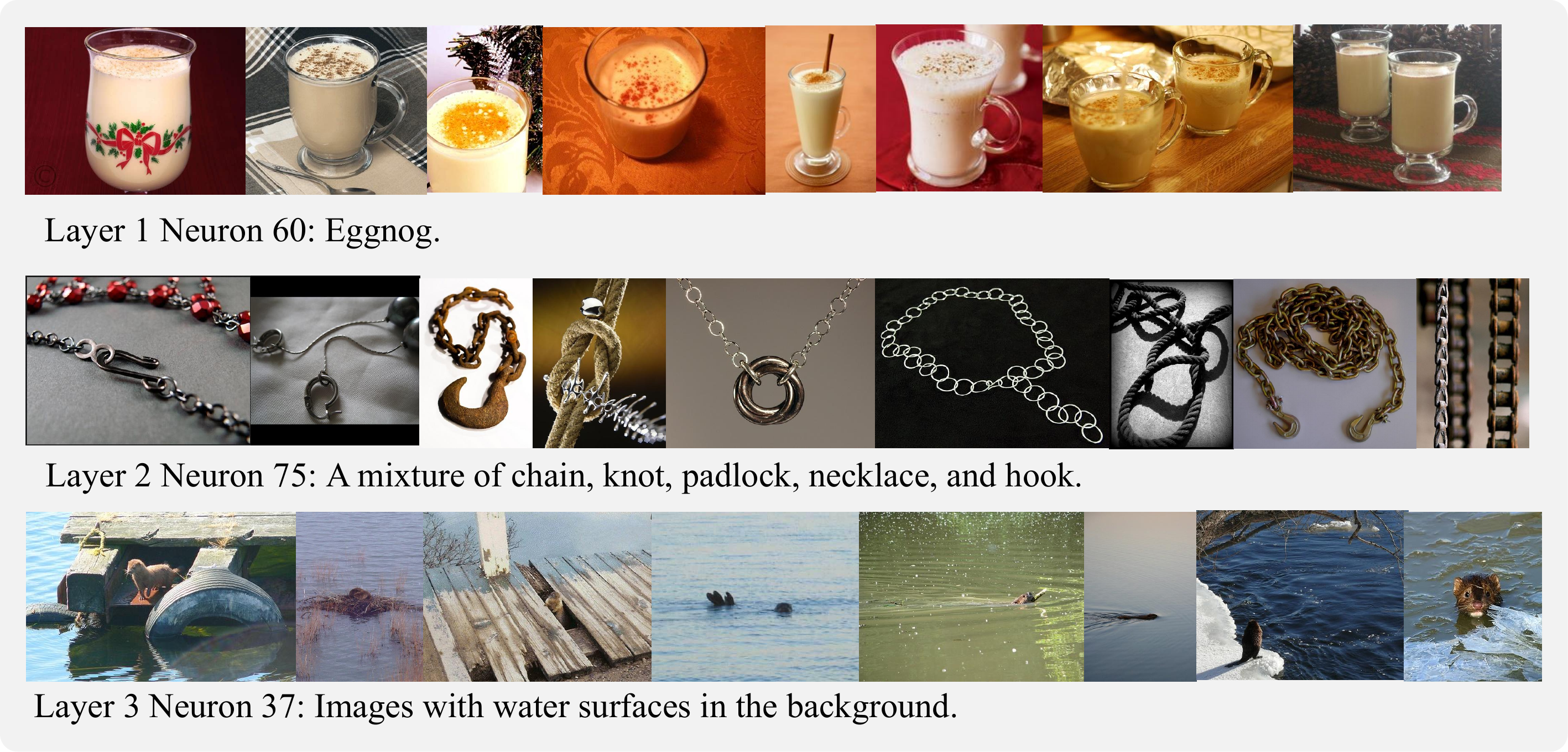}
    \caption{Hierarchical feature discovery in CLIP models: specific objects (Layer 1), functional categories (Layer 2), and abstract background concepts (Layer 3).}
    \label{fig:clip_hierarchy}
\end{figure}

Beyond individual monosemanticity, the core claim of GCFF is the emergence of hierarchical conceptual structure across module layers. Qualitative inspection supports this progression (Figure~\ref{fig:clip_hierarchy}): layer 1 responds to specific, concrete objects (e.g.\ eggnog); layer 2 groups functionally related items (e.g.\ chains, knots, padlocks, necklaces and hooks, which share a fastening or linking function rather than an appearance); and layer 3 captures abstract environmental features that hold independently of the foreground (e.g.\ images with water surfaces in the background). No supervision of abstraction level is provided at any point; the ordering emerges from depth alone. More examples can be found in Appendix A.

\subsection{Classification Performance}
\label{sec:classification}

While GCFF is highly effective as an interpretation module, it is also a performant, standalone forward-forward training algorithm. To demonstrate that biological plausibility does not come at the expense of task performance, we evaluated GCFF on standard image classification benchmarks: MNIST, CIFAR-10, STL-10 and CIFAR-100.
As shown in Table~\ref{tab:classification_results}, GCFF achieves state-of-the-art performance among forward-forward learning algorithms, consistently outperforming the baseline SCFF and original FF algorithms across all evaluated datasets.

\begin{table}[h]
    \centering
    \begin{tabular}{l c c c c}
        \toprule
        \textbf{Method} & \textbf{MNIST} & \textbf{CIFAR-10} & \textbf{STL-10} & \textbf{CIFAR-100} \\
        \midrule
        BackProp  & 98.8\% & 87.9\% & 91.7\% & 65.86\% \\
        \midrule
        FF \citep{hinton2022forwardforwardalgorithmpreliminaryinvestigations} & 94.0\% & 60.6\% & - & - \\
        IntFF \citep{tang2023integratedforwardforwardalgorithmintegrating} & 98.0\% & - & - & - \\
        SCFF \citep{chen2025selfcontrastiveforwardforwardalgorithm} & 98.2\% & 74.9\% & 66.7\% & 48.7\% \\
        \midrule
        \textbf{GCFF (Ours)} & \textbf{98.4\%} & \textbf{75.8\%} & \textbf{68.4\%} & \textbf{49.9\%} \\
        \bottomrule
    \end{tabular}
    \caption{Classification accuracy across standard benchmarks. GCFF achieves state-of-the-art performance among forward-forward training paradigms.}
    \label{tab:classification_results}
\end{table}

\section{Conclusion}
We presented Group-Contrastive Forward-Forward (GCFF), a biologically plausible forward-forward training algorithm that offers an alternative path toward monosemanticity through architectural constraints and contrastive learning rather than sparsity. Applied as an interpretation module on top of the pretrained representations of a vision-language model, a single GCFF module produces monosemantic neurons whose abstraction increases progressively with depth, without any sparsity constraint or explicit supervision of abstraction level. Because its neurons are non-linear functions of the representation under study rather than directions within it, GCFF is not bound by the Linear Representation Hypothesis that constrains sparse dictionary learning. These results suggest that biologically plausible training algorithms hold the potential to produce biologically plausible representations. Beyond interpretability, GCFF also achieves state-of-the-art performance among forward-forward algorithms on image classification benchmarks, demonstrating that biological plausibility and task performance are mutually compatible.

\bibliography{iclr2026_conference}
\bibliographystyle{iclr2026_conference}

\appendix
\newpage
\section*{Appendix A. Qualitative Examples of Identified Visual Features}
\label{sec:appendix}
\vspace{-0.5cm}
\begin{figure}[H]
    \centering
    \includegraphics[width=1\linewidth]{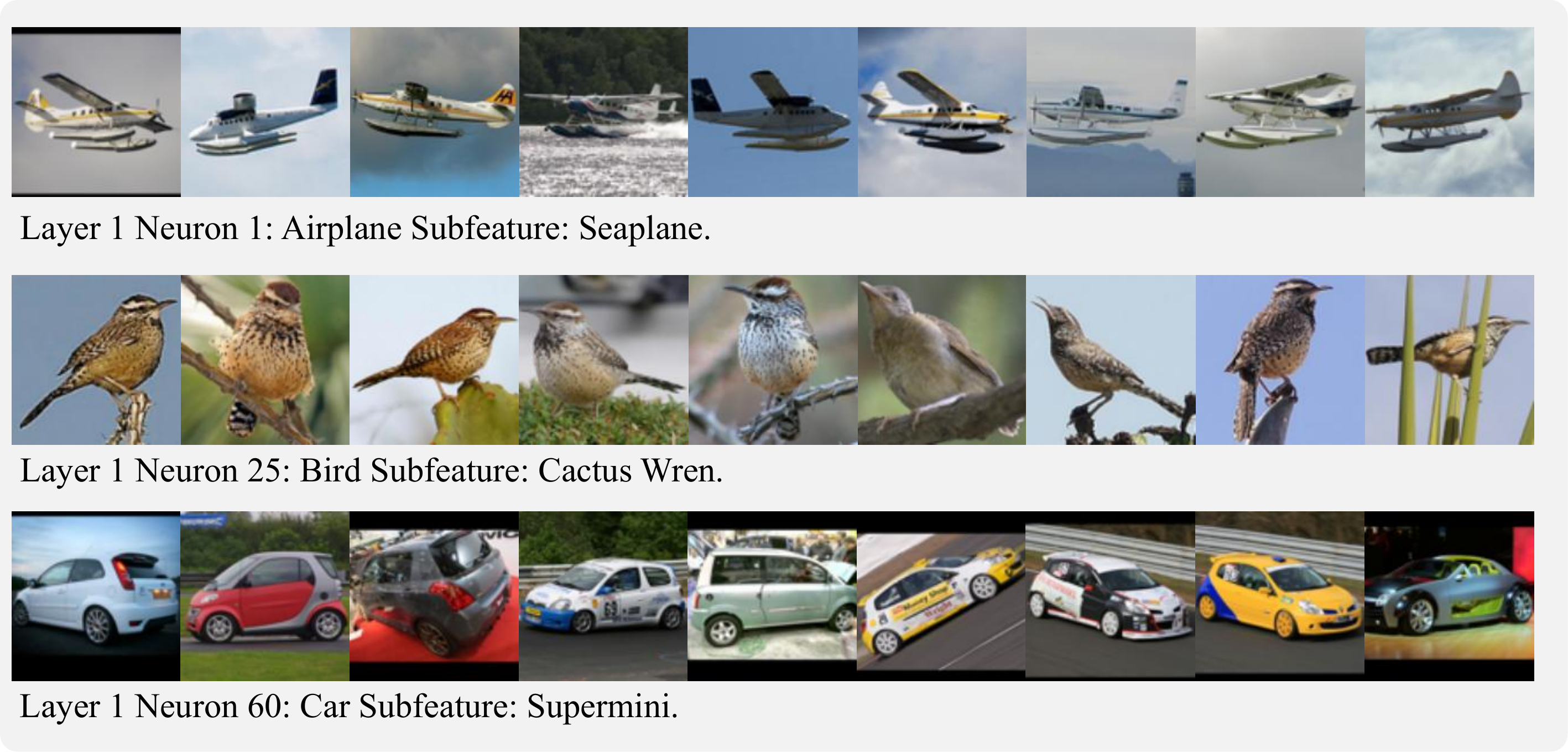}
    \vspace{-0.7cm}
    \caption{Examples of features discovered by GCFF in the first layer. Dataset: STL10.}
    \label{fig:example_layer_1_stl10}
\end{figure}
\vspace{-0.5cm}
\begin{figure}[H]
    \centering
    \includegraphics[width=1\linewidth]{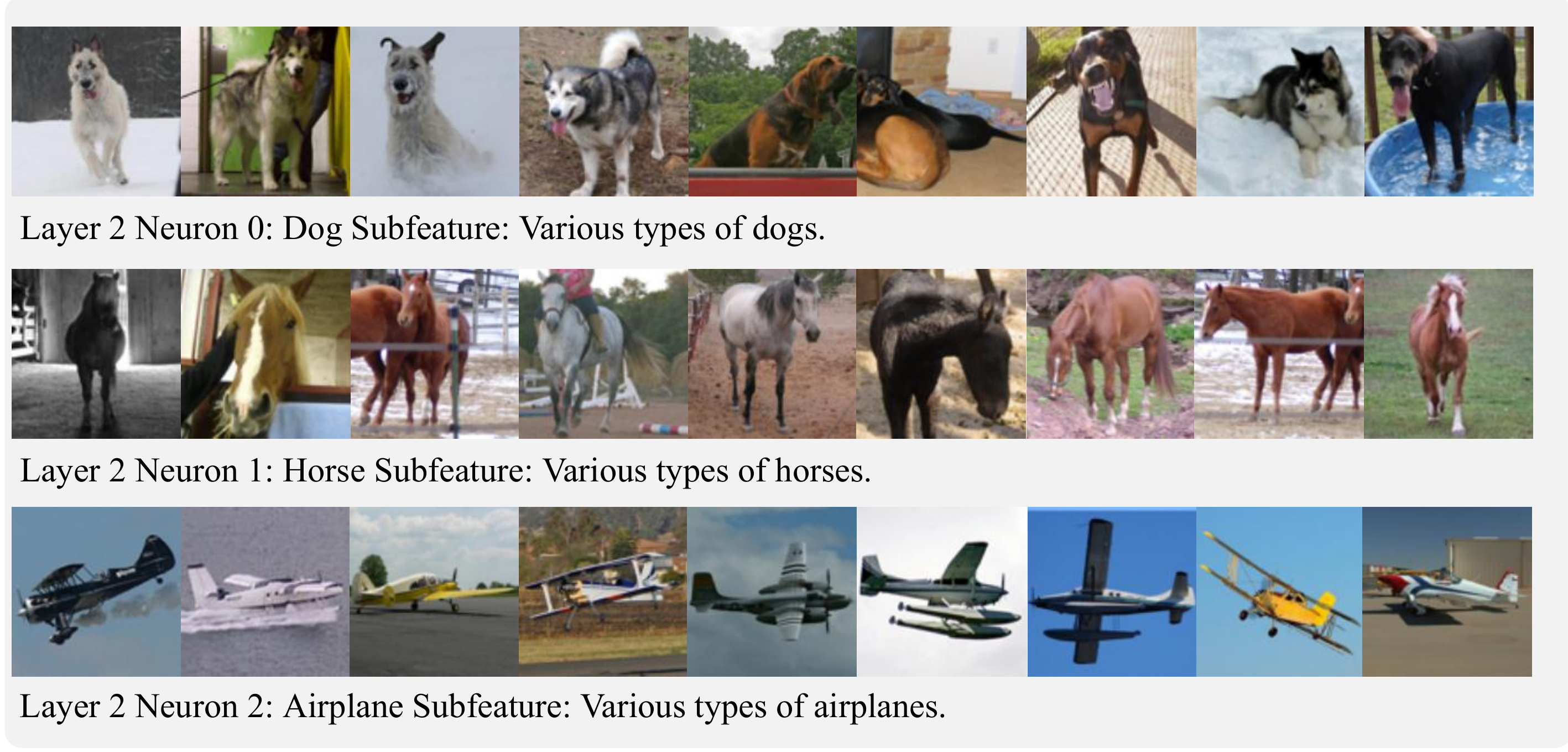}
    \vspace{-0.7cm}
    \caption{Examples of features discovered by GCFF in the second layer. Dataset: STL10.}
    \label{fig:example_layer_2_stl10}
\end{figure}
\vspace{-0.5cm}
\begin{figure}[H]
    \centering
    \includegraphics[width=1\linewidth]{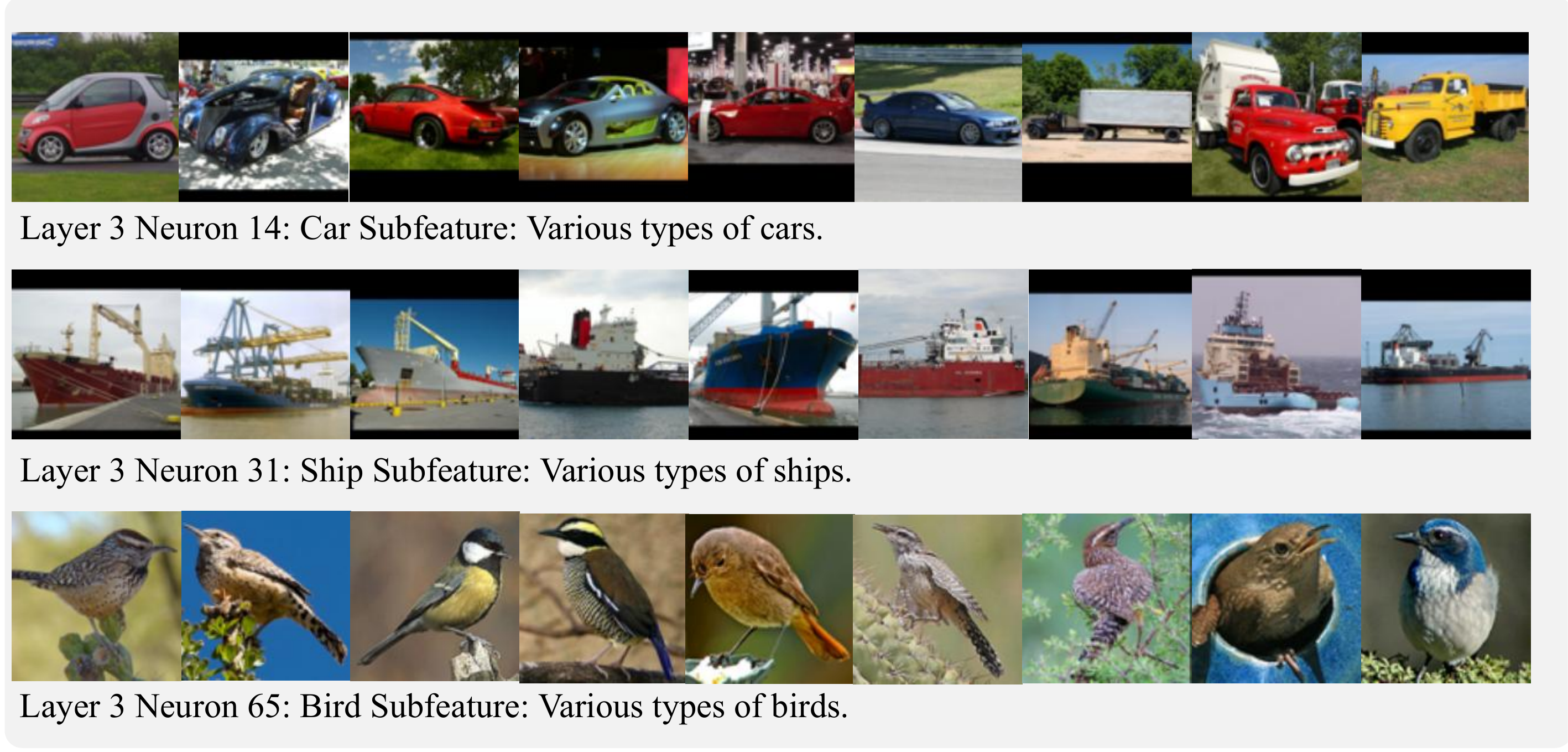}
    \vspace{-0.7cm}
    \caption{Examples of features discovered by GCFF in the third layer. Dataset: STL10.}
    \label{fig:example_layer_3_stl10}
\end{figure}

\newpage
\vspace{-0.3cm}
\begin{figure}[H]
    \centering
    \includegraphics[width=1\linewidth]{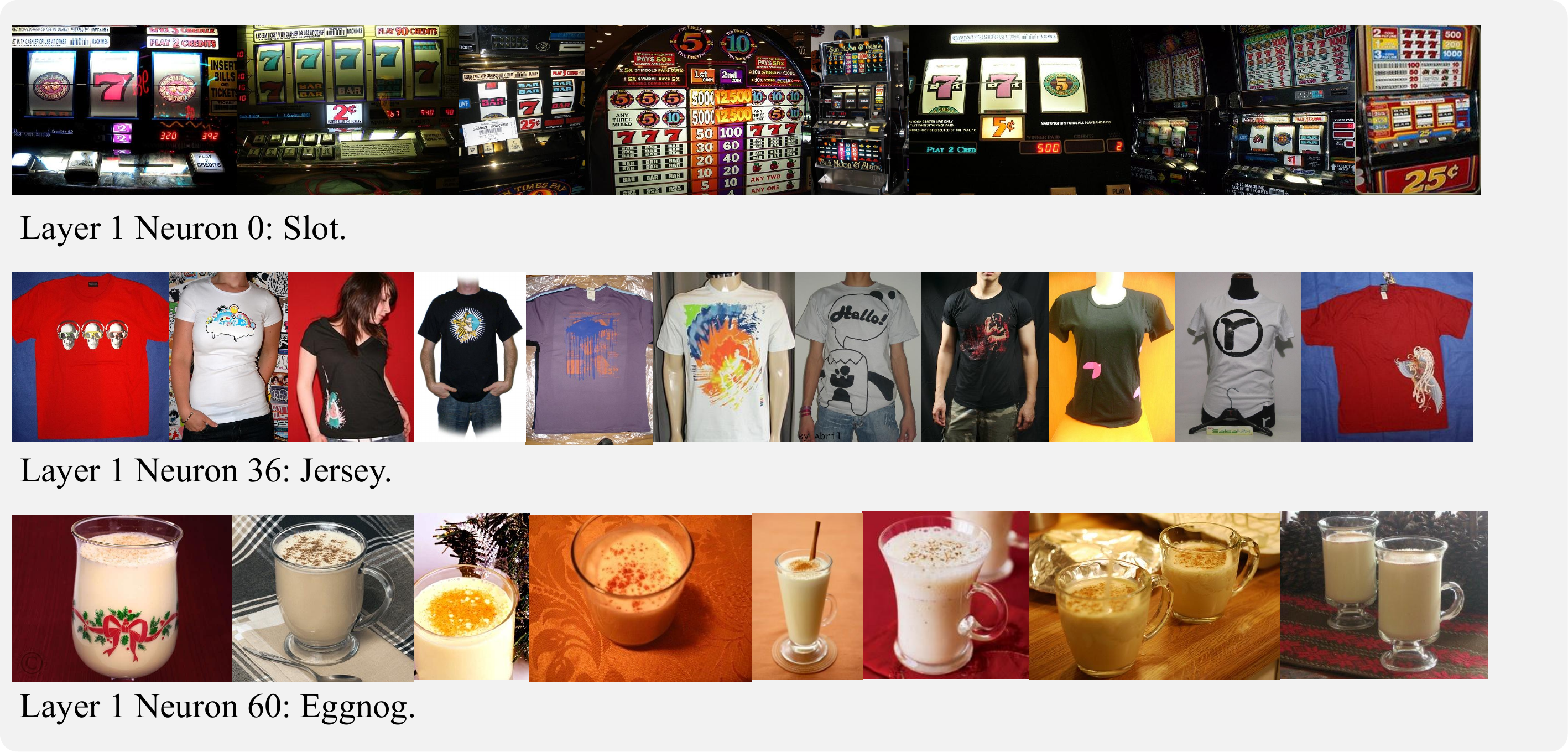}
    \vspace{-0.7cm}
    \caption{Examples of features discovered by GCFF in the first layer. Dataset: Imagenet1k.}
    \label{fig:example_layer_1_imagenet1k}
\end{figure}
\vspace{-0.3cm}
\begin{figure}[H]
    \centering
    \includegraphics[width=1\linewidth]{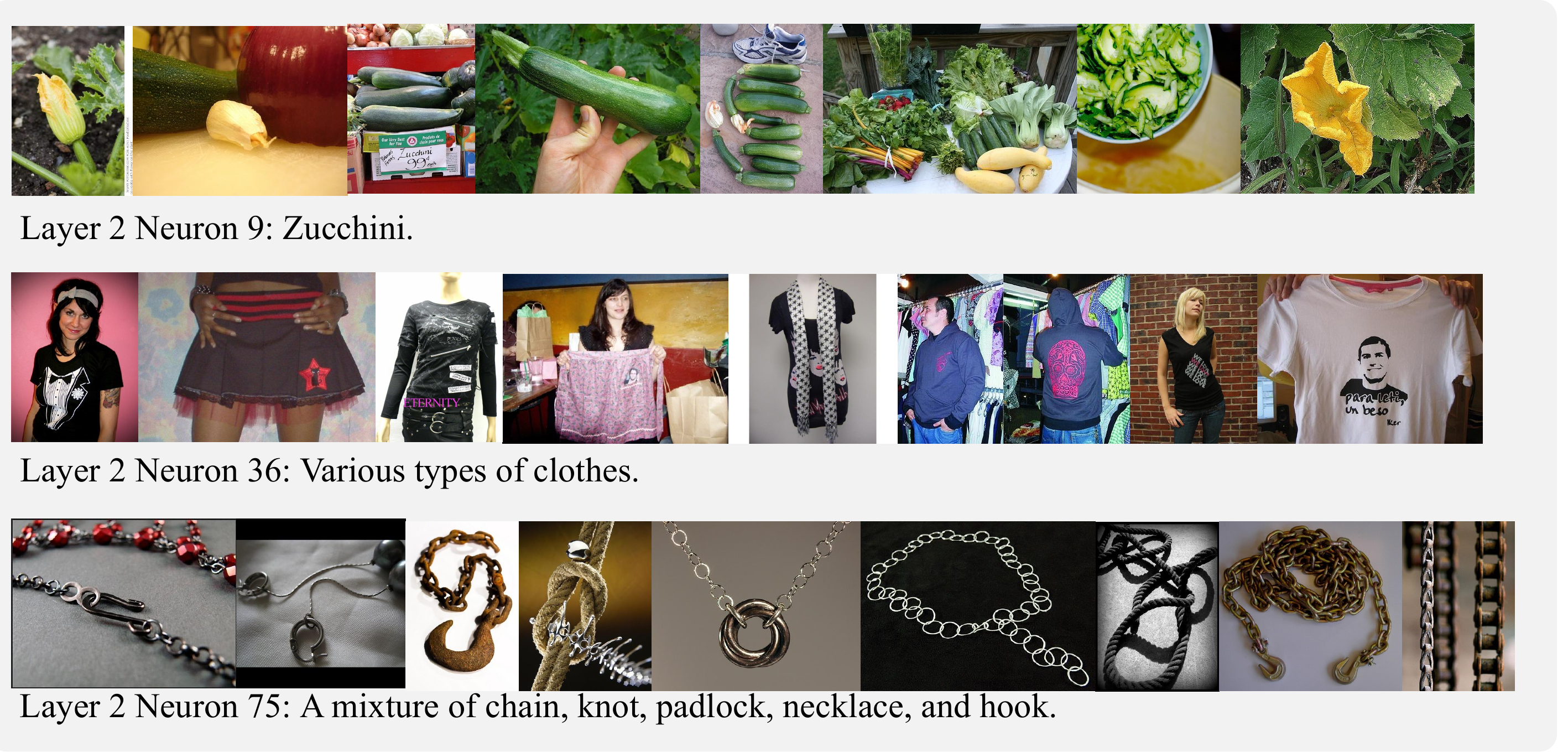}
    \vspace{-0.7cm}
    \caption{Examples of features discovered by GCFF in the second layer. Dataset: Imagenet1k.}
    \label{fig:example_layer_2_imagenet1k}
\end{figure}
\vspace{-0.3cm}
\begin{figure}[H]
    \centering
    \includegraphics[width=1\linewidth]{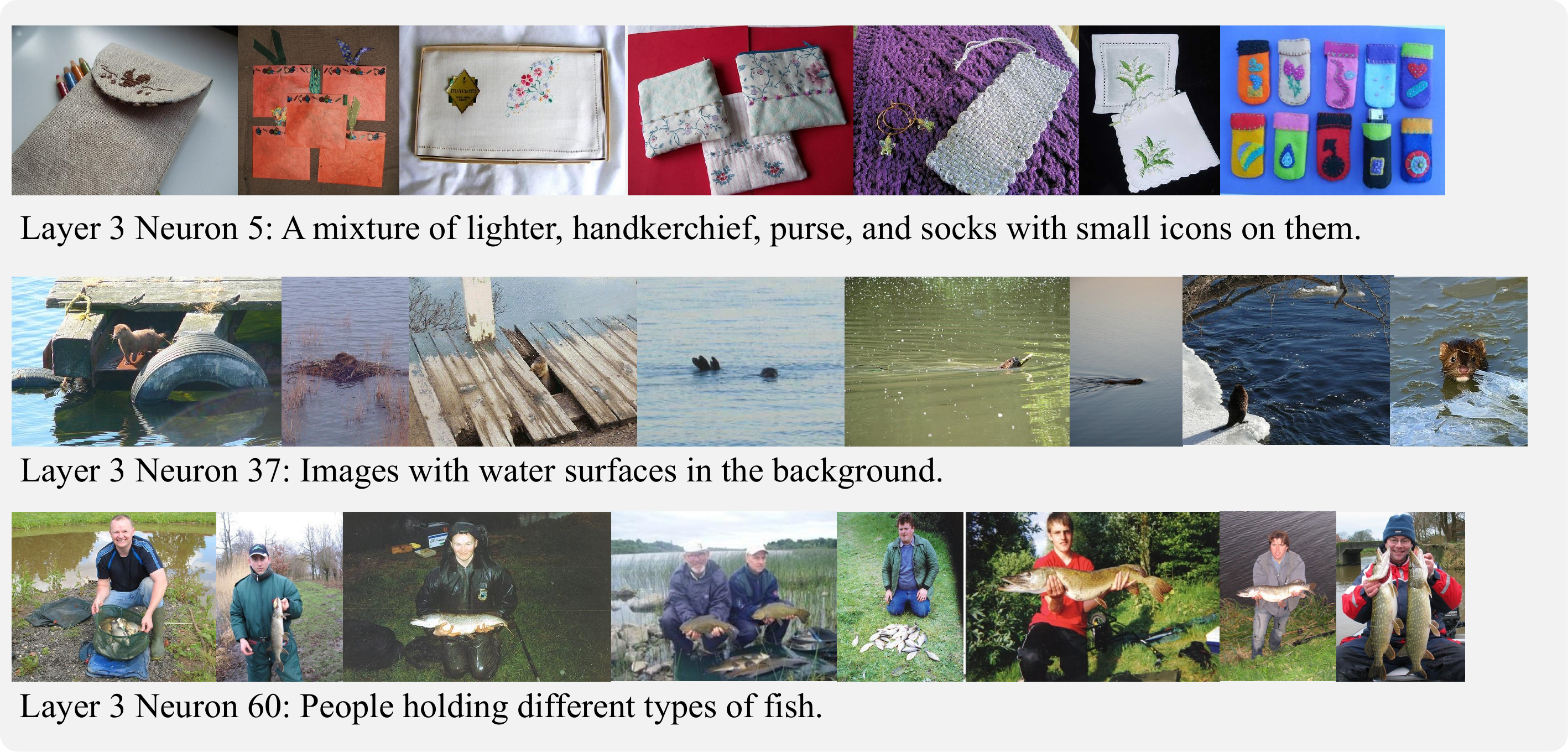}
    \vspace{-0.7cm}
    \caption{Examples of features discovered by GCFF in the third layer. Dataset: Imagenet1k.}
    \label{fig:example_layer_3_imagenet1k}
\end{figure}

\end{document}